\documentclass[letterpaper]{article} 
\usepackage{aaai2026}
\nocopyright
\usepackage{times}  
\usepackage{helvet}  
\usepackage{courier}  
\usepackage[hyphens]{url}  
\usepackage{graphicx} 
\usepackage{natbib}  
\usepackage{caption} 
\usepackage{algorithm}
\usepackage{algorithmic}
\usepackage{enumitem}
\usepackage{booktabs}       
\usepackage{subcaption}     
\usepackage{tcolorbox}      
\tcbuselibrary{breakable}   
\usepackage{enumitem}       
\usepackage{newfloat}
\usepackage{listings}
\DeclareCaptionStyle{ruled}{labelfont=normalfont,labelsep=colon,strut=off} 
\floatstyle{ruled}
\newfloat{listing}{tb}{lst}{}
\floatname{listing}{Listing}
\title{MTA: A Merge-then-Adapt Framework for Personalized \\Large Language Models}

\author{
    Xiaopeng Li\textsuperscript{\rm 1}\textsuperscript{\dag},
    Yuanjin Zheng\textsuperscript{\rm 1, 2}\textsuperscript{\dag},
    Wanyu Wang\textsuperscript{\rm 1},
    Wenlin Zhang\textsuperscript{\rm 1},
    Pengyue Jia\textsuperscript{\rm 1},
    Yiqi Wang\textsuperscript{\rm 1},
    Maolin Wang\textsuperscript{\rm 1},
    Xuetao Wei\textsuperscript{\rm 1},
    Xiangyu Zhao\textsuperscript{\rm 1}\thanks{Corresponding author.}
}

\affiliations{
    \textsuperscript{\rm 1}City University of Hong Kong\\
    \textsuperscript{\rm 2}Zhejiang University\\

}

\usepackage{bibentry}
\usepackage{color, xspace}

\usepackage{amsmath} 
\usepackage{amssymb} 
\usepackage{booktabs}
\usepackage{multirow} 
\usepackage{graphicx}  
\usepackage{subcaption} 

\begin{document}

\maketitle
\renewcommand{\thefootnote}{\fnsymbol{footnote}}
\footnotetext[2]{These authors contributed equally.} 
\renewcommand{\thefootnote}{\arabic{footnote}} 

\begin{abstract}
Personalized Large Language Models (PLLMs) aim to align model outputs with individual user preferences, a crucial capability for user-centric applications. However, the prevalent approach of fine-tuning a separate module for each user faces two major limitations: (1) storage costs scale linearly with the number of users, rendering the method unscalable; and (2) fine-tuning a static model from scratch often yields suboptimal performance for users with sparse data.
To address these challenges, we propose \textbf{MTA}, a Merge-then-Adapt framework for PLLMs. MTA comprises three key stages. First, we construct a shared \emph{Meta-LoRA Bank} by selecting anchor users and pre-training meta-personalization traits within meta-LoRA modules. Second, to ensure scalability and enable dynamic personalization combination beyond static models, we introduce an \emph{Adaptive LoRA Fusion} stage. This stage retrieves and dynamically merges the most relevant anchor meta-LoRAs to synthesize a user-specific one, thereby eliminating the need for user-specific storage and supporting more flexible personalization. Third, we propose a \emph{LoRA Stacking for Few-Shot Personalization} stage, which applies an additional ultra-low-rank, lightweight LoRA module on top of the merged LoRA. Fine-tuning this module enables effective personalization under few-shot settings. Extensive experiments on the LaMP benchmark demonstrate that our approach outperforms existing SOTA methods across multiple tasks. 
\end{abstract}


\section{Introduction}

Large Language Models (LLMs) have demonstrated remarkable capabilities in semantic understanding and text generation across a wide range of tasks. Despite their general success, most current LLMs follow a ``one-size-fits-all'' paradigm during generation, which overlooks user-specific preferences and contextual alignment. This limitation significantly hinders their applicability in domains that inherently require personalized interactions, such as healthcare~\cite{johnson2021precision}, education~\cite{pratama2023revolutionizing}, personalized customer service~\cite{ma2021one, li2025survey}, and recommendation systems~\cite{li2023hamur, gao2024hierrec, zhao2025joint}, etc. Personalized Large Language Models (PLLMs), which tailor their outputs to user-specific preferences, have emerged as a timely and critical research direction for bridging the gap between general-purpose models and individualized user needs.

Early research on PLLMs incorporates explicit user information through prompt-based methods, employing in-context learning~\cite{liu2024once}, retrieval augmentation~\cite{mysore2023pearl}, or profile augmentation~\cite{richardson2023integrating}. However, these approaches are highly sensitive to input noise~\cite{tan2024personalized} and pose potential risks of privacy leakage~\cite{zhang2025proper}. In contrast, recent research has increasingly focused on fine-tuning-based methods, particularly Parameter-Efficient Fine-Tuning (PEFT) methods such as LoRA~\cite{hu2022lora}, which implicitly encode personalization within model parameters. These approaches enable fine-grained alignment to user preferences while mitigating privacy exposure. OPPU~\cite{tan2024democratizing} is the first to introduce a one-LoRA-per-user paradigm, in which user preferences and behavioral patterns are encoded within personal parameters, resulting in superior generation quality. 
Building on this method, Per-Pcs~\cite{tan2024personalized} decomposes complete LoRA modules into smaller components and introduces a dynamic routing mechanism that recombines these elements during inference, thereby improving both personalization adaptability and computational efficiency. PROPER~\cite{wozniak2024personalized} employs a hierarchical modeling framework that operates across population, group, and user levels to enable fine-grained personalized alignment. 

However, these approaches face two key limitations: \textit{(1). Poor Scalability}: The one-LoRA-per-user approach necessitates the independent training and storage of LoRA modules for each user, resulting in parameter growth scaling linearly with the number of users, rendering it impractical for large-scale deployment. Although Per-Pcs partially alleviates this issue through component recombination, it still relies on a static assembly of pre-trained LoRAs and lacks the adaptability necessary to tailor representations to target users.
\textit{(2). Limited performance in sparse-data scenarios.} For OPPU and Per-Pcs, these methods require sufficient training data to achieve parameter convergence and to enable personalized alignment. However, their performance degrades in data-sparse settings, where alignment cannot be reliably established. While PROPER mitigates this issue by incorporating population and group-level training, its multi-level progressive learning pipeline increases procedural complexity and poses challenges for scalability.

To address these challenges, we propose \textbf{MTA}, a three-stage framework employing ``Merge-then-Adapt'' to enable scalable and data-efficient personalization. Firstly, we introduce the module of \textbf{Meta-LoRA Bank Construction}. We pre-train a set of meta-LoRAs by selecting anchor users with highly distinguishable personalization traits. Secondly, we introduce \textbf{Adaptive LoRA Merging}. Based on the similarity between the meta-LoRAs and the target user profile, multiple meta-LoRAs are retrieved. Their parameters are then dynamically merged according to their relevance to the target user. In this way, personalization goes beyond the static one-LoRA-per-user paradigm, achieving scalability through an infinite user-adaptive linear combination mechanism.
Thirdly, we introduce \textbf{LoRA Stacking for Few-shot Personalization}. To enable personalization under sparse-data conditions, an ultra-low-rank LoRA is stacked atop the merged LoRA from the previous stage. During this phase, both the base LLM and the merged LoRA are kept frozen, only the newly added ultra-low-rank LoRA is updated. Owing to its minimal parameter size, this strategy enables efficient and robust adaptation even when training samples are extremely limited.
We conduct comprehensive experiments across LaMP benchmarks~\cite{salemi2023lamp} under five different user personalization settings to validate the effectiveness, scalability, and efficiency of our proposed approach.
Our contributions are summarized as follows:
\begin{itemize}[leftmargin=*]
    \item We present MTA, a novel framework that departs from the conventional static one-LoRA-per-user paradigm. MTA builds a meta-LoRA bank and employs a dynamic adapter-merging mechanism, delivering both scalability and effective personalized adaptation. 
    \item To enhance robustness in sparse-data regimes, a LoRA-Stack training strategy is utilized to enable efficient and stable optimization under data-scarce conditions.
    \item Comprehensive experiments on the LAMP benchmark show that our approach achieves state-of-the-art performance, outperforming existing PEFT methods on PLLMs.
\end{itemize}

\section{Preliminaries}

\paragraph{Problem Formulation.}
Following previous studies on PLLMs~\cite{tan2024democratizing, tan2024personalized}, our primary objective is to generate a user-specific response $\mathbf{r}_u$ for a given user $u$ using a PLLM parameterized by $\Theta$. The response generation is conditioned on the user's query $\mathbf{q}_u$ and their historical behavior data $H_u$.

Our framework reformulates the task by decomposing parameter $\Theta$ into two components. The first component, $\Theta^{\text{merged}}$, represents the collaboratively-informed base model constructed by merging modules from a Meta-LoRA Bank to provide a robust personalized foundation.  The second component, $\Theta^{\text{adapt}}$, is a lightweight, low-rank adapter trained on individual user data to capture fine-grained preferences, with the final personalized model integrating both components as follows:
\begin{equation}
\label{eq:task_formulation_revised}
\mathbf{r}_u = \text{LLM}(\mathbf{q}_u ; \Theta^{\text{merged}} \oplus \Theta^{\text{adapt}})
\end{equation}
where $\oplus$ denotes the additive combination of parameters such that the collaborative knowledge from $\Theta^{\text{merged}}$ is augmented with user-specific adaptations from $\Theta^{\text{adapt}}$, enabling both broad personalization capabilities and individual fine-tuning within a unified architecture.

\begin{figure*}[t]
    \centering
    \includegraphics[width=\textwidth, page=1]{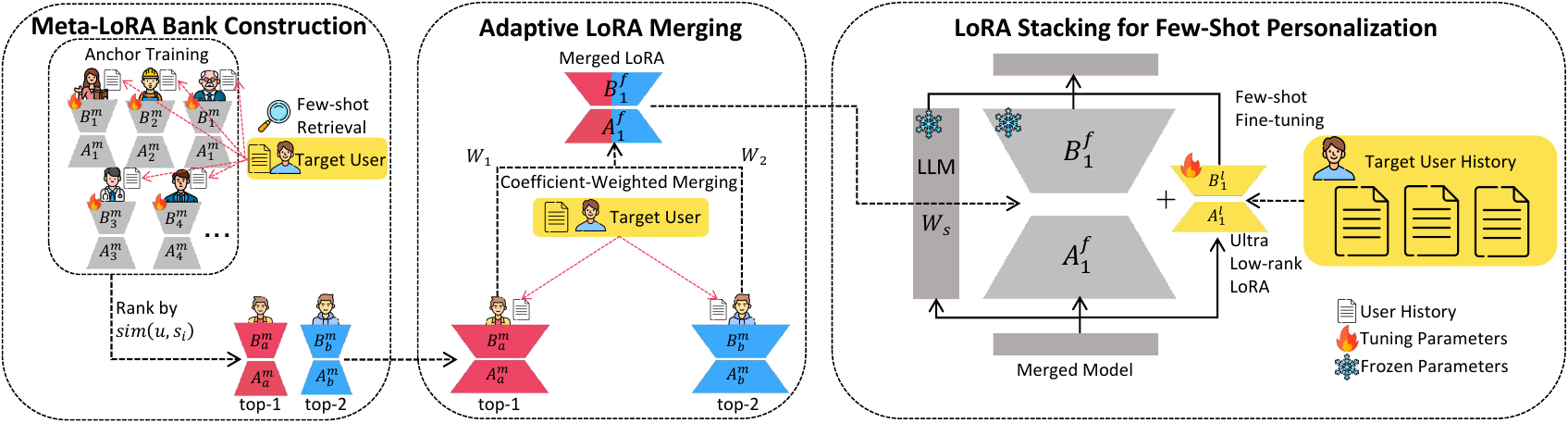}
    \caption{An overview of the MTA (Merge-then-Adapt) framework is as follows: The \textbf{left} panel, Meta-LoRA Bank Construction, shows the pre-training of a bank of anchor modules and the retrieval of the top two most relevant LoRAs for a target user. The \textbf{middle} panel, Adaptive LoRA Merging, shows the two retrieved LoRAs being combined through a coefficient-weighted merge to create a single, personalized LoRA. The \textbf{right} panel, LoRA Stacking for Few-Shot Personalization, depicts the freezing of the merged model and the fine-tuning of a new, ultra low-rank LoRA on top of it using the user's history for final adaptation.}
    \label{fig:meta-lora-wide}
\end{figure*}
\section{Methodology}

In this section, we provide an overview of our framework and detail each of its stages.

\subsection{Overall Framework}

To overcome the critical limitations of poor scalability and ineffectiveness in data-sparse settings, we propose \textbf{MTA}, a Merge-then-Adapt framework, as shown in Figure~\ref{fig:meta-lora-wide}. MTA consists of a three stages process:

\begin{itemize}[leftmargin=*]
    \item \textbf{Stage 1 (Meta-LoRA Bank Construction):} To build a bank of foundational modules with diverse personalization traits, we first generate embeddings for all users from their histories and partition them using K-Means clustering. The most active user from each cluster is selected as an ``anchor user'', and a unique LoRA module is fine-tuned on each one. This collection of anchor modules constitutes the Meta-LoRA Bank.

    \item \textbf{Stage 2 (Adaptive LoRA Merging):} To achieve scalability and move beyond the static one-LoRA-per-user paradigm, for a target user, we retrieve the most similar anchor LoRAs from the bank based on cosine similarity. Their corresponding modules are then combined using a weighted linear merge—with weights directly proportional to their similarity scores—to create a single, customized ``Merged LoRA''.

    \item \textbf{Stage 3 (LoRA Stacking for Few-Shot Personalization):} To enable robust and efficient adaptation in data-scarce settings, we follow a sequential \textbf{Merge-then-Adapt} strategy, the ``Merged LoRA'' is first irreversibly combined with the base model's weights. This entire model is then frozen. Finally, a new, extremely low-rank ``stacked LoRA'' is trained on top using the user's few-shot data, enabling a final, highly-efficient adaptation.
\end{itemize}

\subsection{Meta-LoRA Bank Construction.} 

The conventional one-LoRA-per-user paradigm exhibits $\mathcal{O}(n)$ computational complexity, necessitating linear scaling of both storage and training costs as user populations expand. To address this, we construct a fixed bank of $\mathcal{V}$ ``experts'' LoRAs with diverse yet representative personalization traits, distilled from the broader user population. This approach serves as the foundation for our MTA framework, where $\mathcal{V} \ll n$, effectively reducing computational complexity to $\mathcal{O}(\mathcal{V})$, while ensuring each expert represents a distinct preference profile. This first stage, therefore, focuses on constructing a compact yet representative Meta-LoRA Bank.
This process begins by mapping each user's behavioral history, $H_u = \{h_1, ..., h_M\}$, into a dense vector space. We use a pre-trained \textsc{DeBERTa-v3-Large}~\cite{he2021debertav3} encoder to generate a corresponding embedding $\mathbf{e}_i$ for each history item $h_i$. The user's global preference embedding, $\boldsymbol{E}_u$, is then calculated as the average of all their item embeddings:
\begin{equation}
\label{eq:user_embedding}
\boldsymbol{E}_u = \frac{1}{|H_u|} \sum_{\boldsymbol{h}_i \in H_u} \text{Encoder}(\boldsymbol{h}_i)
\end{equation}

After obtaining embeddings for all candidate users, we employ the K-Means algorithm to partition them into $\mathcal{V}$ disjoint clusters. To ensure each anchor user serves as a high-quality representative of their cluster's preferences, from each cluster $\mathcal{C}_v$, we select the most ``active'' user—the one with the most extensive behavior history—as the anchor user $u_v^*$:
\begin{equation}
\label{eq:anchor_selection}
u_v^* = \underset{u \in \mathcal{C}_v}{\arg\max} \, |H_u|
\end{equation}

Then, we pre-train a dedicated LoRA module for each selected anchor user using their historical data. This collection of modules, $\{\boldsymbol{\Theta}_{u_1^*}, ..., \boldsymbol{\Theta}_{u_\mathcal{V}^*}\}$, constitutes our Meta-LoRA Bank to facilitate the following merging steps.

\subsection{Adaptive LoRA Merging.} 

After constructing the LoRA banks, maintaining the traditional rigid one-to-one mapping between users and LoRA modules would severely limit personalization capabilities and render the approach impractical for large-scale deployment. We propose \textbf{Adaptive LoRA Merging}, which dynamically combines parameters from multiple anchor LoRAs retrieved from the bank based on their relevance to the target user. This approach creates a personalized foundation that implements an infinite user-adaptive linear combination mechanism, enabling a small, fixed set of foundational modules to serve a virtually unlimited number of users.

The theoretical foundation for this approach rests on recent research demonstrating that skills and traits learned by different LoRA modules can be effectively combined through linear operations~\cite{huang2024lorahubefficientcrosstaskgeneralization, zhao2024mergingloraslikeplaying, prabhakar2024lorasoupsmergingloras}. These studies show that linear combinations facilitate generalization to new tasks while enabling the composition of novel abilities, providing the principled basis for our merging strategy.

Our implementation follows a two-step process. First, we identify the most relevant anchor users through embedding-based similarity matching. We concatenate each user's complete text history $H_u$ into a unified document and encode it using the dense embedding model \textsc{BGE-small-en-v1.5}~\cite{bge_embedding} to produce a comprehensive representation $\boldsymbol{E}_u^{\text{retrieval}}$. The relevance between target user $u$ and anchor user $s_i$ is quantified through cosine similarity:

\begin{equation}
\label{eq:cosine_similarity}
\text{sim}(u, s_i) = \frac{\boldsymbol{E}_u^{\text{retrieval}} \cdot \boldsymbol{E}_{s_i}^{\text{retrieval}}}{\|\boldsymbol{E}_u^{\text{retrieval}}\| \|\boldsymbol{E}_{s_i}^{\text{retrieval}}\|}
\end{equation}

Based on these scores, we select the the two most similar anchors, $s_a$ and $s_b$, and perform a weighted linear combination of their LoRA parameters to form the merged LoRA $\boldsymbol{\Theta}_u^{\text{merged}}$:
\begin{equation}
\label{eq:merging}
\boldsymbol{\Theta}_u^{\text{merged}} = \alpha_u^a \cdot \boldsymbol{\Theta}_{s_a} + \alpha_u^b \cdot \boldsymbol{\Theta}_{s_b}
\end{equation}

The merging coefficients, $\alpha_u^a$ and $\alpha_u^b$, are computed by normalizing the similarity scores of their respective anchors. This ensures that the weight assigned to each anchor is directly proportional to its relevance to the target user:
\begin{align}
\label{eq:alpha_calculation_corrected}
\alpha_u^a &= \frac{\text{sim}(u, s_a)}{\text{sim}(u, s_a) + \text{sim}(u, s_b)} \\
\alpha_u^b &= \frac{\text{sim}(u, s_b)}{\text{sim}(u, s_a) + \text{sim}(u, s_b)}
\end{align}

This weighting scheme guarantees that the most similar anchor contributes more significantly to the final merged LoRA, creating a knowledge-rich and highly relevant starting point for the final personalization stage.

\subsection{LoRA Stacking for Few-Shot Personalization.} 

Following the merging step, although the merged module achieves personalization gains, it still lacks fine-grained training on target users. Moreover, sparse data distribution across different users often prevents reliable alignment due to insufficient data for parameter convergence. This necessitates achieving fine-grained training even in data-sparse scenarios. To address this, we introduce \textbf{LoRA Stacking}, a sequential Merge-then-Adapt strategy designed for data efficiency. Instead of continuing to train the larger merged LoRA, we freeze it and train \textit{only} a new, ultra low-rank adaptation adapter on the user's few-shot data. Given its minimal parameter size, this approach enables efficient and robust adaptation even when training samples are extremely limited, ensuring a strong final alignment. 

The first step involves integrating the merged LoRA $\boldsymbol{\Theta}_u^{\text{merged}}$ into the base model weights. This operation creates a new, intermediate model with updated weights $\boldsymbol{W}_{\text{merged}} = \boldsymbol{W}_0 + \boldsymbol{W}_{\text{LoRA}}^{\text{merged}}$, where the collaborative knowledge from anchor users becomes embedded within the model architecture. The resulting merged model is then frozen to serve as an enhanced foundation for subsequent personalization.

The second step introduces a user-specific adaptation layer designed for efficient fine-tuning on limited data. We deploy an ultra low-rank LoRA module, termed the adaptation adapter $\boldsymbol{\Theta}_u^{\text{adapt}}$, which operates on top of the frozen merged foundation. The deliberately minimal parameter count of this adapter enables rapid convergence while maintaining personalization effectiveness. The forward pass for this stacked architecture follows:
\begin{equation}
\label{eq:stacking}
\begin{aligned}
    \boldsymbol{h} &= \boldsymbol{W}_{\text{merged}}\boldsymbol{x} + \boldsymbol{W}_u^{\text{adapt}}\boldsymbol{x} \\
         & = (\boldsymbol{W}_0 + \boldsymbol{W}_{\text{LoRA}}^{\text{merged}})\boldsymbol{x} + \boldsymbol{B}_u^{\text{adapt}}\boldsymbol{A}_u^{\text{adapt}}\boldsymbol{x}
\end{aligned}
\end{equation}
During training, the optimization process focuses exclusively on the parameters of $\Theta_u^{\text{adapt}}$ while minimizing loss on the user's limited historical data $H_u$. The adaptation adapter employs an ultra-low-rank configuration (e.g., $r=4$), which substantially reduces trainable parameters and ensures computationally efficient convergence. This design enables effective personalization even when user data is severely limited, making the approach particularly suitable for real-world deployment scenarios where extensive user histories are unavailable.

\begin{table*}[!t]
\centering
\small 
\setlength{\tabcolsep}{4pt} 
\begin{tabular}{@{}llccccccc@{}}
\toprule
\multirow{2}{*}{\textbf{Task}} & \multirow{2}{*}{\textbf{Metric}} & \multicolumn{3}{c}{\textbf{Prompt-based (RAG)}} & \multicolumn{4}{c}{\textbf{Fine-tuning-based}} \\ \cmidrule(lr){3-5} \cmidrule(lr){6-9}
 & & $k$=1 & $k$=3 & $k$=5 & OPPU & PER-PCS & PROPER & \textbf{MTA(Ours)} \\ \midrule
\multirow{2}{*}{\parbox{4.5cm}{LaMP-1: PERSONALIZED\\CITATION IDENTIFICATION}} & Acc ↑ & .4915 & .5085 & .4407 & .5085 & \underline{.5339} & .5169 & \textbf{.5424$^*$} \\
 & F1 ↑ & .4821 & .4962 & .4087 & .5037 & \underline{.5289} & .5110 & \textbf{.5412$^*$} \\ \midrule
\multirow{2}{*}{\parbox{4.5cm}{LaMP-2: PERSONALIZED\\MOVIE TAGGING}} & Acc ↑ & .2917 & .3472 & .4028 & .4167 & .3611 & \underline{.4306} & \textbf{.4444$^*$} \\
 & F1 ↑ & .2054 & .2510 & \underline{.3131} & .2368 & .1911 & \textbf{.3161$^*$} & .2592 \\ \midrule
\multirow{2}{*}{\parbox{4.5cm}{LaMP-3: PERSONALIZED\\PRODUCT RATING}} & MAE ↓ & .3964 & .4144 & .4144 & .2432 & .2162 & \underline{.2072} & \textbf{.1982$^*$} \\
 & RMSE ↓ & .7229 & .8383 & .8490 & .5452 & .5199 & \underline{.4746} & \textbf{.4350$^*$} \\ \midrule
\multirow{2}{*}{\parbox{4.5cm}{LaMP-4: PERSONALIZED\\NEWS HEADLINE GEN.}} & R-1 ↑ & .1664 & .1928 & .1890 & .2144 & .1891 & \underline{.2152} & \textbf{.2399$^*$} \\
 & R-L ↑ & .1332 & .1611 & .1621 & \underline{.1885} & .1821 & .1849 & \textbf{.2134$^*$} \\ \midrule
\multirow{2}{*}{\parbox{4.5cm}{LaMP-5: PERSONALIZED\\SCHOLARLY TITLE GEN.}} & R-1 ↑ & .4663 & .4926 & .4887 & .4836 & .4958 & \underline{.5122} & \textbf{.5291$^*$} \\
 & R-L ↑ & .3839 & .4184 & .4141 & .4254 & .4370 & \underline{.4487} & \textbf{.4659$^*$} \\ \bottomrule
\end{tabular}%
\caption{Comparison of MTA against baselines on the LaMP benchmark. For the RAG method, $k$ denotes the number of retrieved user history items. The metrics R-1 and R-L refer to ROUGE-1 and ROUGE-L, respectively. The $\uparrow$ arrow indicates that higher values are better, while the $\downarrow$ arrow indicates that lower values are preferable. The best-performing result for each metric is highlighted in \textbf{bold} and the second-best is \underline{underlined}. An asterisk (*) indicates statistically significant improvements (i.e., a two-sided t-test with $p<0.05$) over the best baseline.}
\label{tab:main_results}
\end{table*}
\section{Experiment}
\subsection{Experimental Setting}
\paragraph{Datasets.}
Following previous work~\cite{tan2024democratizing, tan2024personalized,zhang-etal-2024-star}, we conduct experiments on the Large Language Model Personalization (LaMP) benchmark~\cite{salemi2023lamp}, which features seven public tasks spanning four classification and three generation scenarios( see Appendix~\S A for task details)\footnote{We exclude the LaMP-6 task due to restricted access to its private dataset. Additionally, we exclude LaMP-7 task as its user history lacks the query-answer correspondence required to construct a training dataset for our framework.}. A key focus of our work is to address the challenge of personalization in data-scarce scenarios. Therefore, following previous works~\cite{tan2024democratizing,tan2024personalized}, for the test set of each task, we randomly select 100 users from the benchmark who possess a notably limited behavioral history. The remaining users are utilized for the construction of our Meta-LoRA Bank. This strict separation ensures that the users in the test set and those contributing to the bank are completely disjoint. Further details regarding the dataset statistics are available in the Appendix~\S B.

\paragraph{Baselines.}
We compare our proposed framework against two main categories of personalization baselines: prompt-based and fine-tuning-based methods. For the prompt-based baseline, we adopt the Retrieval-Augmented Personalization (RAG) method proposed in LaMP~\cite{salemi2023lamp}, which augments the user's query with the top-k most relevant items retrieved from their personal history corpus. For the more advanced fine-tuning-based methods, we compare against several state-of-the-art frameworks, including OPPU~\cite{tan2024democratizing}, PER-PCS~\cite{tan2024personalized}, and PROPER~\cite{wozniak2024personalized}. Further details on the implementation of each baseline can be found in the Appendix~\S C.

\paragraph{Evaluation Metrics.}
Our evaluation protocol strictly adheres to the standards established by the LaMP benchmark~\cite{salemi2023lamp}. For the text classification tasks (LaMP-1, LaMP-2), we assess performance using \textbf{accuracy} and \textbf{F1-score}. The personalized product rating task (LaMP-3), which can be treated as a regression problem, is evaluated using \textbf{Mean Absolute Error (MAE)} and \textbf{Root Mean Squared Error (RMSE)}. For all text generation tasks (LaMP-4, LaMP-5), we measure output quality using \textbf{ROUGE-1} and \textbf{ROUGE-L}~\cite{lin-2004-rouge}. 

\paragraph{Experimental Details.}
Detailed hyperparameter settings and the specific prompt templates used in our experiments are provided in Appendix \S D and \S E. To ensure a fair and powerful comparison, we use \textsc{Meta-Llama-3-8B-Instruct}~\cite{grattafiori2024llama3herdmodels} as the foundational backbone LLM for all evaluated approaches, including our own. All experiments are conducted on a server equipped with a single NVIDIA L20 (48GB) GPU and 20 vCPUs of an Intel(R) Xeon(R) Platinum 8457C processor.

\subsection{Overall Performance}
Table~\ref{tab:main_results} shows the performance of our proposed framework and baselines on the curated test sets across five tasks from the LaMP benchmark. We have the following observations:

\noindent\textbf{MTA vs. RAG.} As shown in Table~\ref{tab:main_results}, our proposed framework, MTA, shows a clear performance advantage over the Retrieval-Augmented Generation (RAG) baseline in the vast majority of tasks and metrics. For instance, MTA achieves relative gains of 6.67\% in accuracy and 9.07\% in F1-score for the citation identification task (LaMP-1) when compared against RAG's best performance across different k-values. This trend is even more pronounced in the product rating prediction task (LaMP-3), where our method yields substantial improvements of 50.00\% in MAE and 39.83\% in RMSE. For news headline generation (LaMP-4), our framework also shows relative improvements of 24.43\% in ROUGE-1 and 31.65\% in ROUGE-L. These results highlight the benefits of our hybrid merging and stacking approach over relying solely on retrieval for LLM personalization.

\noindent\textbf{MTA vs. OPPU.} Our framework's ``merge-then-adapt'' strategy shows a distinct advantage over the OPPU baseline, which trains a LoRA module from scratch and is less effective in data-sparse settings. The empirical results in Table \ref{tab:main_results} confirm this. For instance, in the movie tagging task (LaMP-2), our method achieves relative gains of 6.65\% in accuracy and 9.46\% in F1-score. This performance improvement is also evident in the news headline generation task (LaMP-4), where our framework secures relative improvements of 11.89\% in ROUGE-1 and 13.21\% in ROUGE-L. These results underscore the superiority of our approach, which leverages collaborative knowledge to create a more robust and data-efficient starting point for personalization.

\noindent\textbf{MTA vs. PER-PCS.} While both our framework and \textbf{PER-PCS} leverage collaborative knowledge, our ``merge-then-adapt'' strategy demonstrates superior performance by incorporating a final, adaptive training stage. PER-PCS assembles a personalized model from pre-existing LoRA ``pieces'' without retraining, which can limit its ability to capture highly unique user profiles. The results in Table ~\ref{tab:main_results} validate this advantage. For example, in the movie tagging task (LaMP-2), MTA achieves relative improvements of 23.07\% in accuracy and 35.64\% in F1-score over PER-PCS. This performance gap is also substantial in the product rating task (LaMP-3), where our method obtains relative gains of 8.33\% in MAE and 16.33\% in RMSE. These results show that while assembling from shared components is effective, the final low-rank training step in our MTA is crucial to learn the fine-grained, residual preferences of individual users.

\noindent\textbf{MTA vs. PROPER.} Our framework also compares favorably to PROPER, a more complex, progressive training baseline. While PROPER utilizes a sophisticated Mixture-of-Experts and dual-router system, our simpler ``merge-then-adapt'' approach achieves competitive or superior results with significantly less architectural complexity, as demonstrated in Table \ref{tab:main_results}. For instance, in the scholarly title generation task (LaMP-5), MTA surpasses PROPER with relative gains of 3.30\% in ROUGE-1 and 3.83\% in ROUGE-L. A similar advantage is observed in the product rating task (LaMP-3), where our method achieves relative improvements of 4.34\% in MAE and 8.34\% in RMSE. These findings suggest that our direct fusion and targeted residual adaptation strategy offers a more efficient path to strong personalization, avoiding the overhead associated with training complex routing and expert mechanisms.

\subsection{Ablation Experiment}
\begin{table}[!t]
\centering
\setlength{\tabcolsep}{0.5pt}
\begin{tabular}{@{}llccc@{}}
\toprule
\textbf{Task} & \textbf{Metric} & \parbox{2cm}{\centering\textbf{Adapt-Only\\LoRA}} & \parbox{2.2cm}{\centering\textbf{Merged-Only\\LoRA}} & \textbf{MTA (Ours)} \\ \midrule
\multirow{2}{*}{LaMP-1} & ~~Acc $\uparrow$ & .5254 & \underline{.5339} & \textbf{.5424} \\
                        & ~~F1 $\uparrow$  & .5182 & \underline{.5323} & \textbf{.5412} \\ \midrule
\multirow{2}{*}{LaMP-2} & ~~Acc $\uparrow$ & .2917 & \underline{.3611} & \textbf{.4444} \\
                        & ~~F1 $\uparrow$  & \underline{.2054} & .2031 & \textbf{.2592} \\ \midrule
\multirow{2}{*}{LaMP-3} & ~~MAE $\downarrow$ & .3514 & \underline{.2342} & \textbf{.1982} \\
                        & ~~RMSE $\downarrow$& .8329 & \underline{.4840} & \textbf{.4350} \\ \midrule
\multirow{2}{*}{LaMP-4} & ~~R-1 $\uparrow$  & .1757 & \underline{.2374} & \textbf{.2399} \\
                        & ~~R-L $\uparrow$  & .1467 & \underline{.2127} & \textbf{.2134} \\ \midrule
\multirow{2}{*}{LaMP-5} & ~~R-1 $\uparrow$  & .4885 & \underline{.5043} & \textbf{.5291} \\
                        & ~~R-L $\uparrow$  & .4058 & \underline{.4525} & \textbf{.4659} \\ \bottomrule
\end{tabular}
\caption{Ablation study comparing our complete framework against its core components. The best performance for each metric is in \textbf{bold}, and the second-best is \underline{underlined}.}
\label{tab:ablation}
\end{table}
To validate the effectiveness and synergy of each key component within our proposed framework, we conduct a detailed ablation study with the results presented in Table~\ref{tab:ablation}. This study compares our complete framework, MTA, against the following two ablated variants:
\begin{itemize}[leftmargin=*]
    \item \textbf{Adapt-Only LoRA}: This variant bypasses the merging stage entirely. It directly fine-tunes a new, extremely low-rank LoRA on the base LLM using only the user's few-shot data. This isolates the effect of the final ``adapt'' step without the benefit of our collaborative warm-start.
    \item \textbf{Merged-Only LoRA}: This variant performs only the adaptive LoRA merging, omitting the final adaptation step. It isolates the effect of the collaborative warm-start.
\end{itemize}

The results in Table~\ref{tab:ablation} clearly demonstrate that our complete framework (MTA) consistently and significantly outperforms both ablated versions. The performance of ``Merged-Only LoRA'' is substantially better than that of ``Adapt-Only LoRA'', which confirms that starting from a collaboratively-informed, merged base is far more effective than fine-tuning from a generic one. Furthermore, the significant performance jump from ``Merged-Only LoRA'' to our full MTA framework across all tasks highlights that the final, low-rank adaptation step is crucial for capturing the fine-grained, specific preferences of the user. In conclusion, both the initial merge and the final adaptation are integral and synergistic components of our framework's success.

\subsection{Efficiency Experiment}
\begin{figure}[t]
    \centering
    \includegraphics[width=0.40\textwidth]{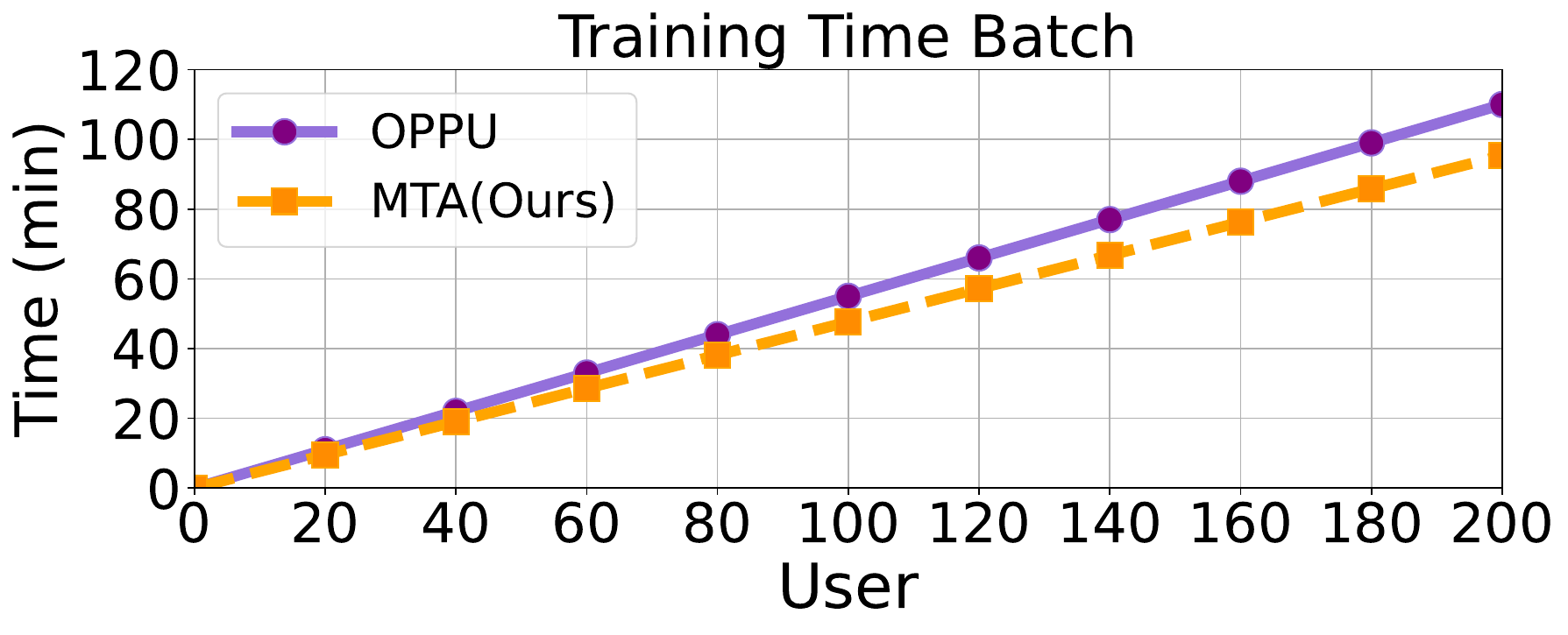}
    \includegraphics[width=0.40\textwidth]{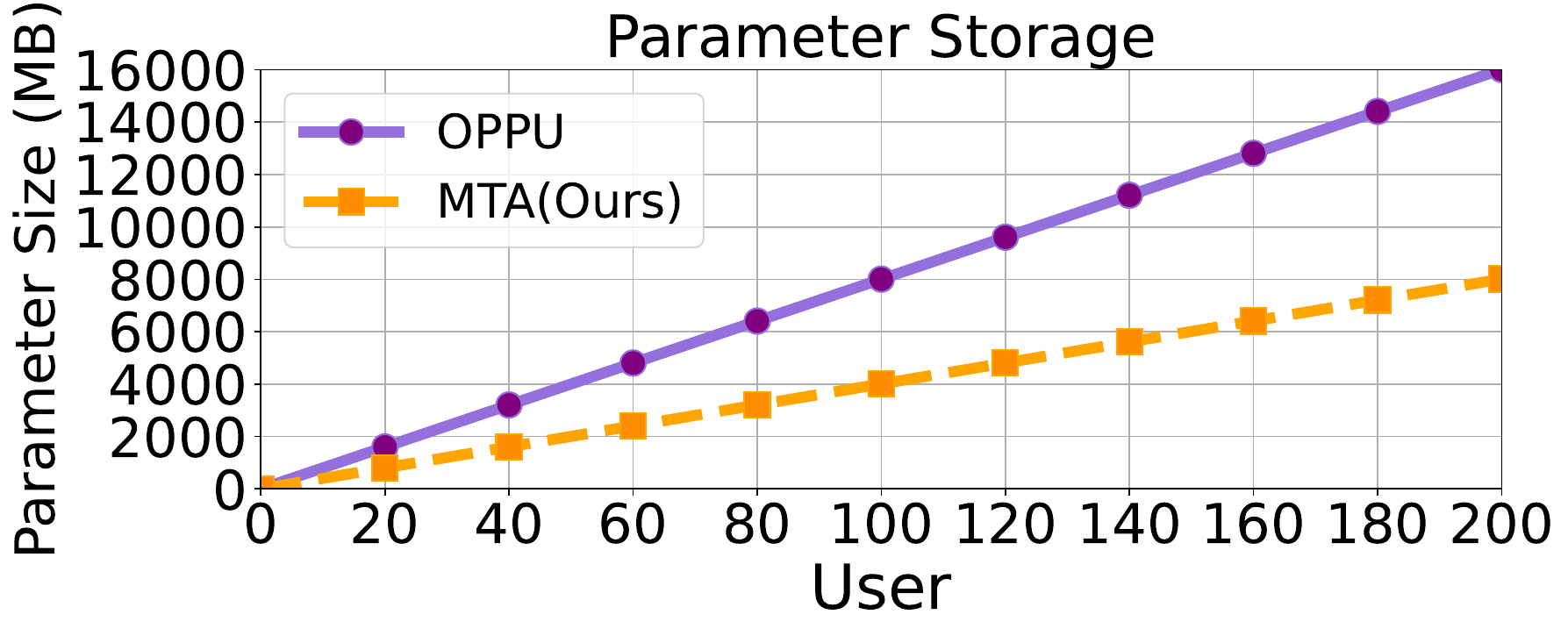}
    \caption{Efficiency comparison between our MTA framework and the OPPU baseline. The top plot shows total training time versus the number of users; the bottom plot shows total parameter storage.}
    \label{fig:efficiency_analysis}
\end{figure}

We evaluate the efficiency of our framework in terms of average total training time and parameter storage across five benchmark tasks, comparing MTA against the OPPU baseline. As shown in Figure~\ref{fig:efficiency_analysis}, MTA demonstrates significant advantages in both metrics. The efficiency stems from only needing to store a single, ultra-low-rank adapter per user and leveraging a strong merged foundation for faster convergence. While a training-free method like PER-PCS provides superior inference speed, it lacks a crucial adaptation stage for the target user, leading to degraded performance as validated in our experiments. Conversely, the hierarchical PROPER method incurs substantial costs: its initial stages demand exceptionally high parameter storage, and its multi-stage training paradigm incurs significant computational costs. These comparisons confirm that MTA offers a more balanced and scalable solution for large-scale deployment.

\subsection{Parameter Experiment}
In this section, we explore three key parameters that affect our results: the Merging Coefficient, which determines the proportional influence of each selected anchor LoRA; the Number of Merged Anchors, which sets how many modules are combined; and the Adaptation LoRA Rank, which controls the parameter budget for the final personalization stage (detailed in Appendix~\S F).

\paragraph{Analysis of the Merging Coefficient $\alpha_u$.}
In our framework, the merging coefficient $\alpha_u$ is adaptively determined based on user similarity. To validate the effectiveness of this adaptive approach, we conduct an ablation study comparing it against several fixed merging coefficients. To demonstrate this, we select one representative task from both classification and generation domains: LaMP-3 (Personalized Product Rating) and LaMP-5 (Personalized Scholarly Title Generation), respectively. We test several fixed values for the merging coefficient, which are sampled at equal intervals between 0 and 1.

\begin{figure}[!t]
    \centering
    \begin{subfigure}[b]{0.49\columnwidth}
        \includegraphics[width=\textwidth]{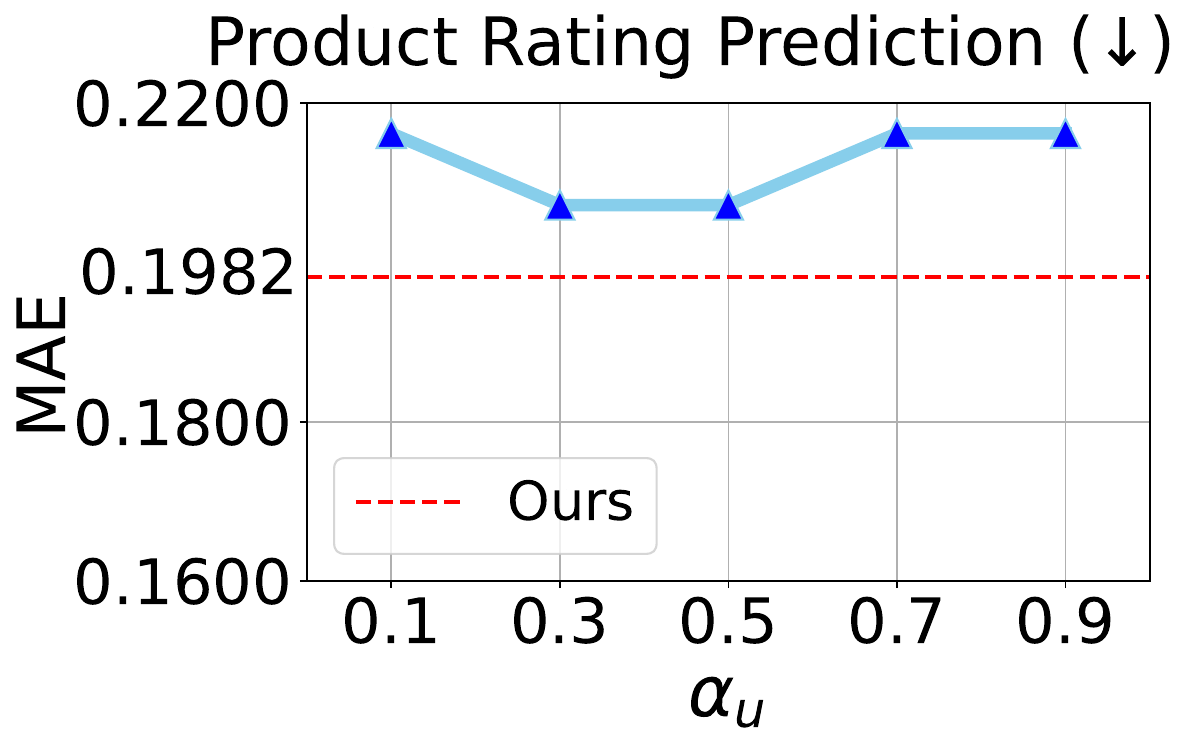}
    \end{subfigure}
    \hfill 
    \begin{subfigure}[b]{0.49\columnwidth}
        \includegraphics[width=\textwidth]{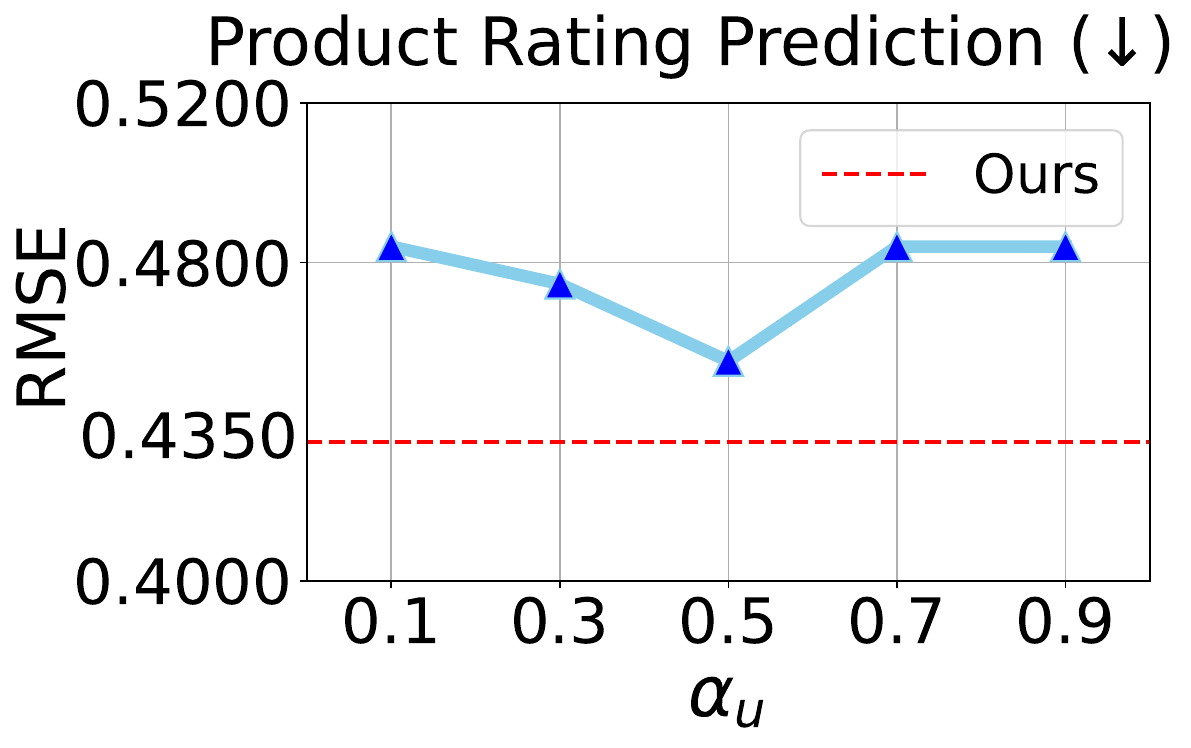}
    \end{subfigure}
    
    
    \begin{subfigure}[b]{0.49\columnwidth}
        \includegraphics[width=\textwidth]{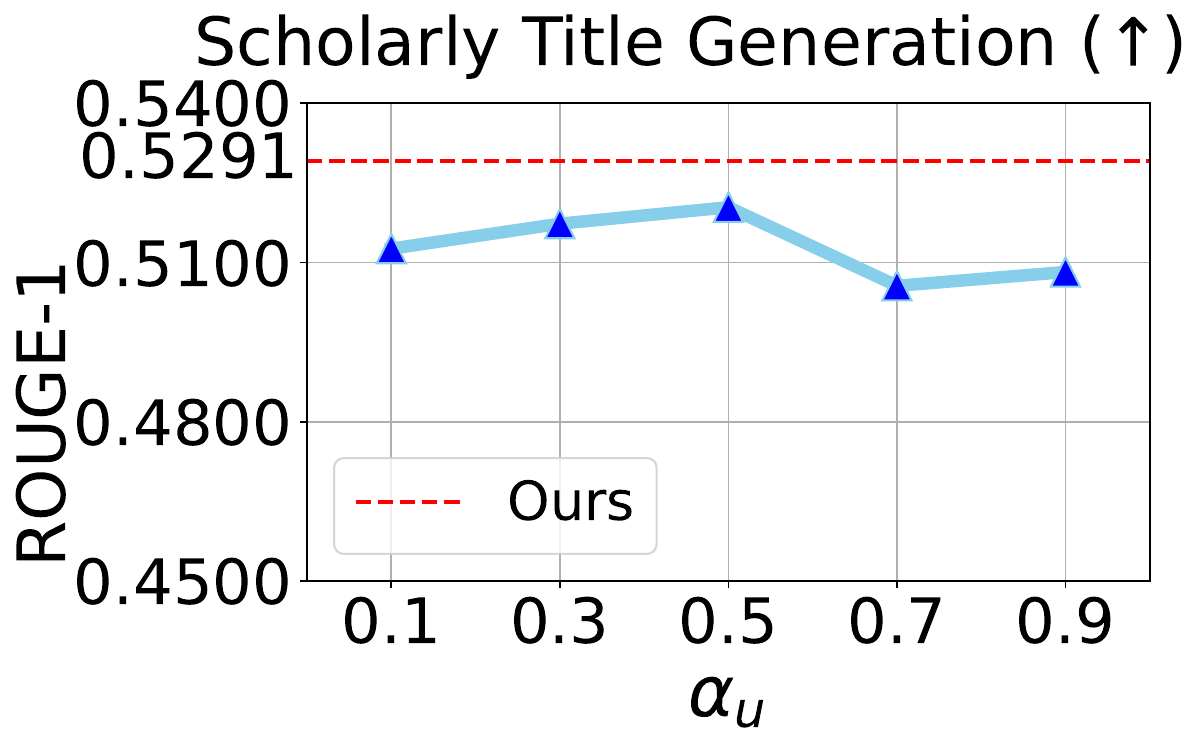}
    \end{subfigure}
    \hfill
    \begin{subfigure}[b]{0.49\columnwidth}
        \includegraphics[width=\textwidth]{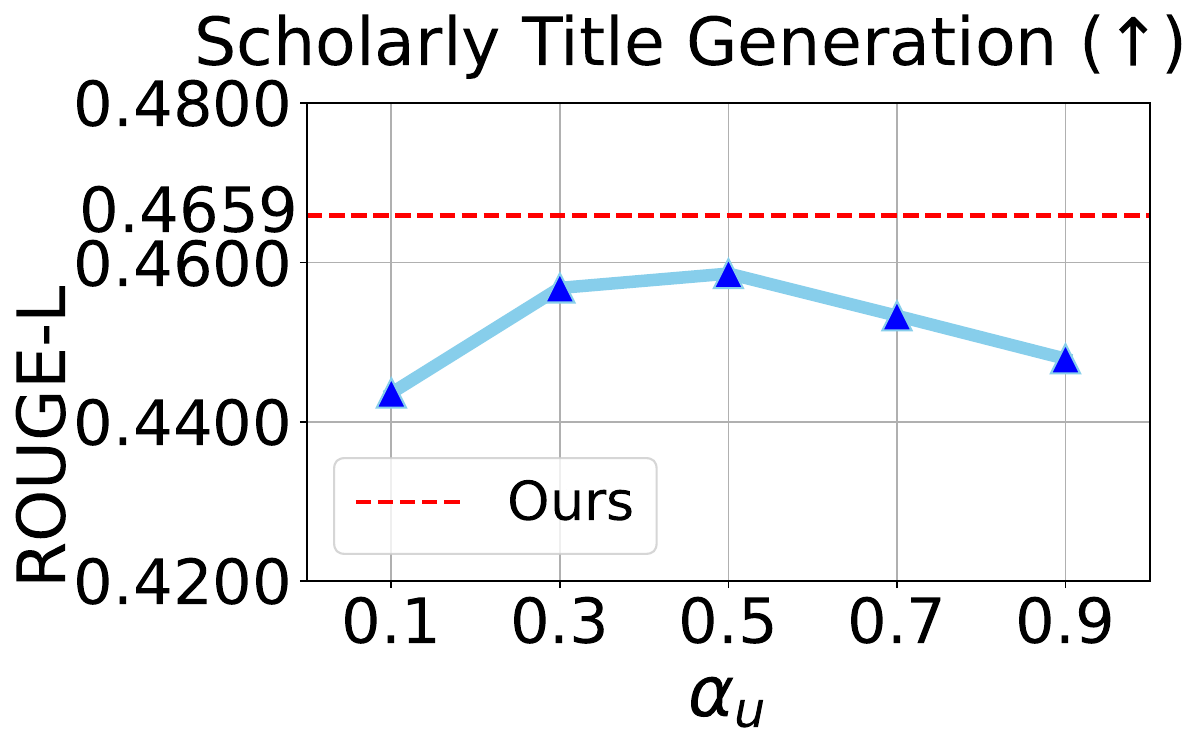}
    \end{subfigure}
    
    \caption{Performance comparison on Personalized Product Rating Prediction (LaMP-3) (top row, MAE/RMSE, lower is better) and Personalized Scholarly Title Generation (LaMP-5) (bottom row, ROUGE-1/L, higher is better) tasks with different fixed merging coefficients ($\alpha_u$) versus our adaptive method (red dashed line). The blue solid line shows the performance of fixed-alpha variants.}
    \label{fig:alpha_ablation}
\end{figure}

As illustrated in Figure~\ref{fig:alpha_ablation}, our adaptive method consistently outperforms all fixed-alpha settings across every tested metric. This demonstrates the clear superiority of our similarity-based adaptive merging approach.
\paragraph{Impact of the Number of Merged Anchors (Top-K).}
To examine the impact of the number of merged anchors on performance, we conduct an analysis where we vary this number $K$ from 2 to 8. For each $K$ value, the top-$K$ most similar anchor users $\{s_1, ..., s_K\}$ are selected, and their LoRA modules are merged. The merged weights are normalized based on their similarity scores, generalizing the merging process as follows:
\begin{equation}
\label{eq:generalized_merging}
\boldsymbol{\Theta}_u^{\text{merged}} = \sum_{i=1}^{K} \alpha_i \cdot \boldsymbol{\Theta}_{s_i}, \quad \left( \alpha_i = \frac{\text{sim}(u, s_i)}{\sum_{j=1}^{K} \text{sim}(u, s_j)} \right)
\end{equation}
We evaluate this on two representative tasks: LaMP-3 (Product Rating) and LaMP-5 (Scholarly Title Generation).

For the Product Rating task (LaMP-3), the best performance is achieved at $K$=3. While a larger $K$ can be beneficial for certain tasks, it also adds extra computing cost as the cost of merging scales linearly with the number of anchors. Conversely, for the Scholarly Title Generation task (LaMP-5), performance is optimal at $K$=2, and including more anchors beyond this point leads to performance degradation, likely due to noise from less-relevant peers.
\begin{figure}[!h]
    \centering
    \begin{subfigure}[b]{0.49\columnwidth}
        \includegraphics[width=\textwidth]{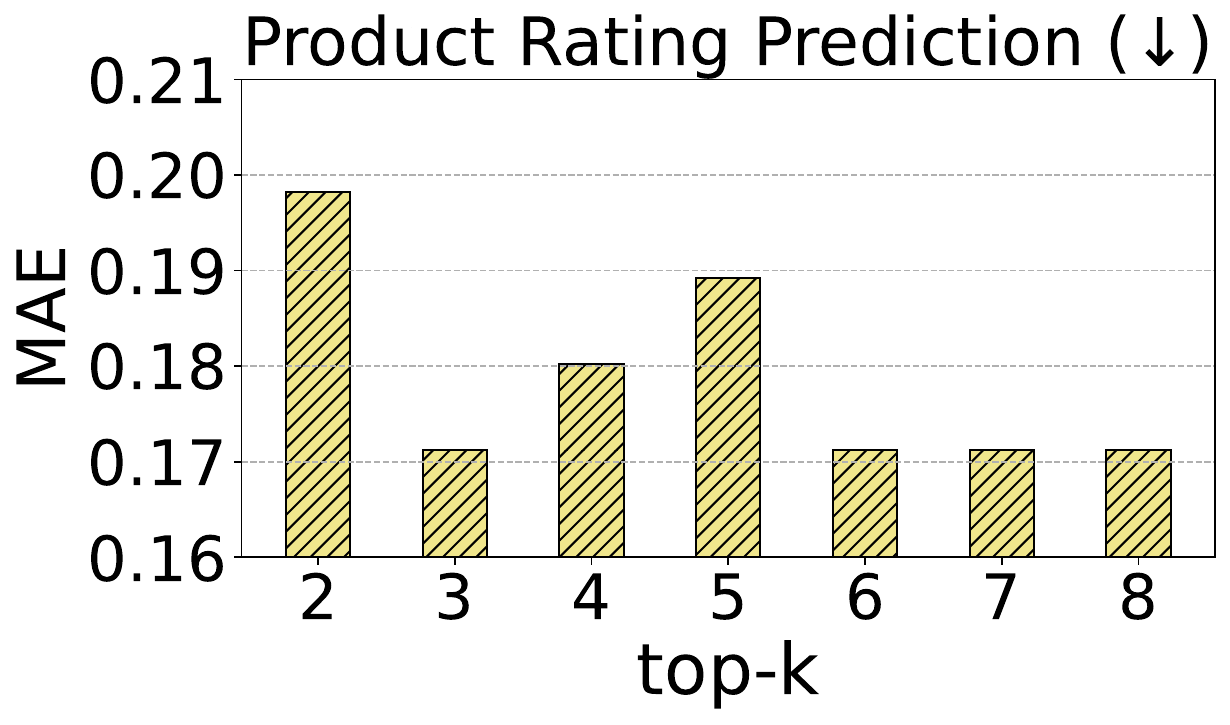}
    \end{subfigure}
    \hfill
    \begin{subfigure}[b]{0.49\columnwidth}
        \includegraphics[width=\textwidth]{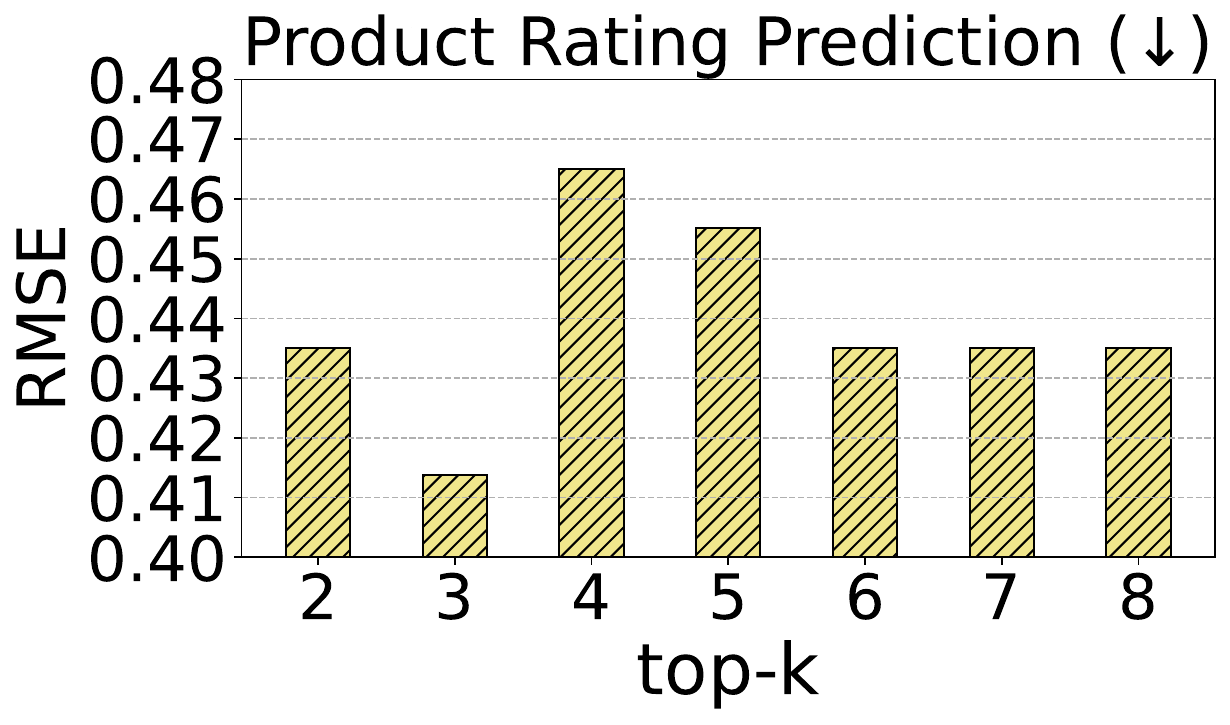}
    \end{subfigure}

    
    \begin{subfigure}[b]{0.49\columnwidth}
        \includegraphics[width=\textwidth]{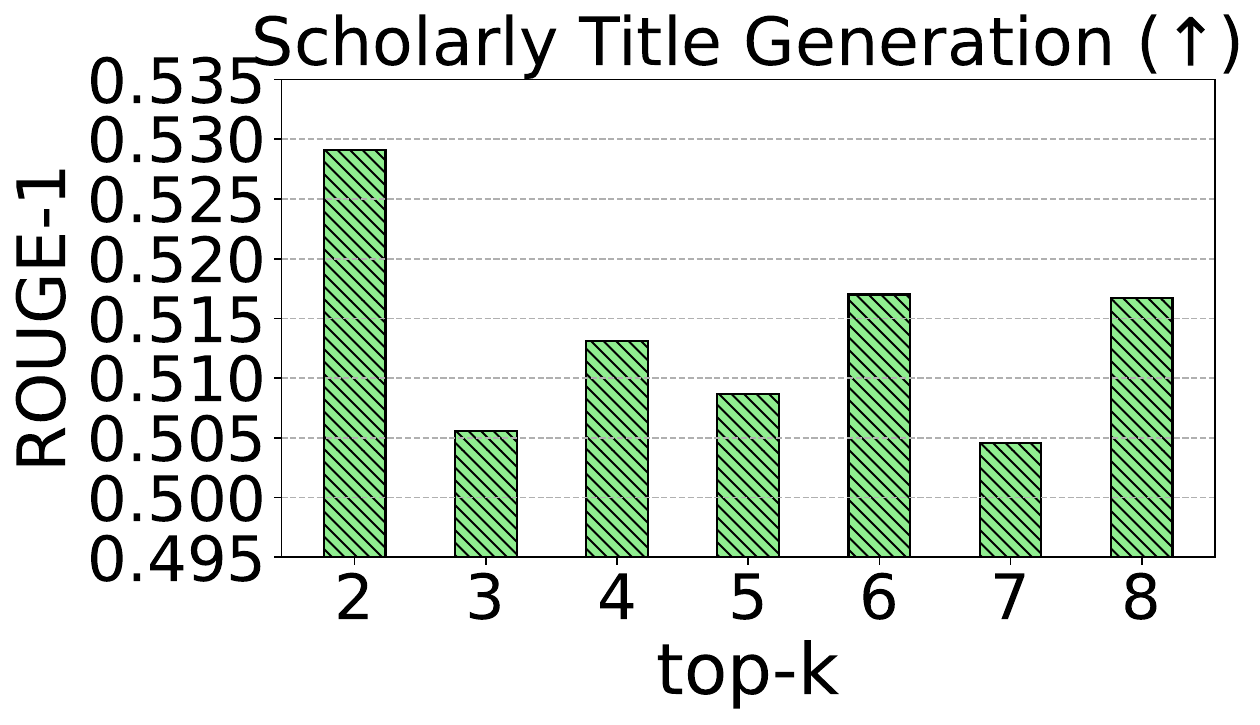}
    \end{subfigure}
    \hfill
    \begin{subfigure}[b]{0.49\columnwidth}
        \includegraphics[width=\textwidth]{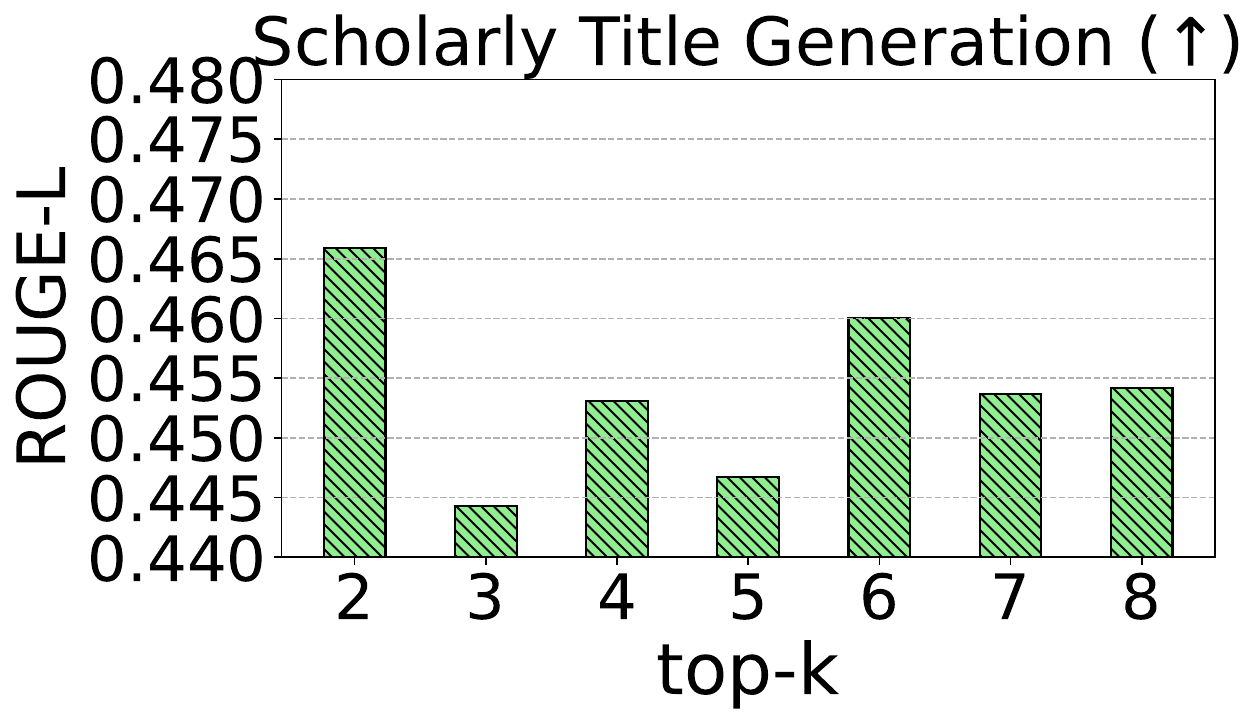}
    \end{subfigure}
    
\caption{Performance on Personalized Product Rating Prediction (LaMP-3) (top row, MAE/RMSE, lower is better) and Personalized Scholarly Title Generation (LaMP-5) (bottom row, ROUGE-1/L, higher is better) as a function of the number of merged top-K anchor users.}
    \label{fig:topk_analysis}
\end{figure}

\begin{figure}[t]
    \centering
    \includegraphics[width=\columnwidth, page=2]{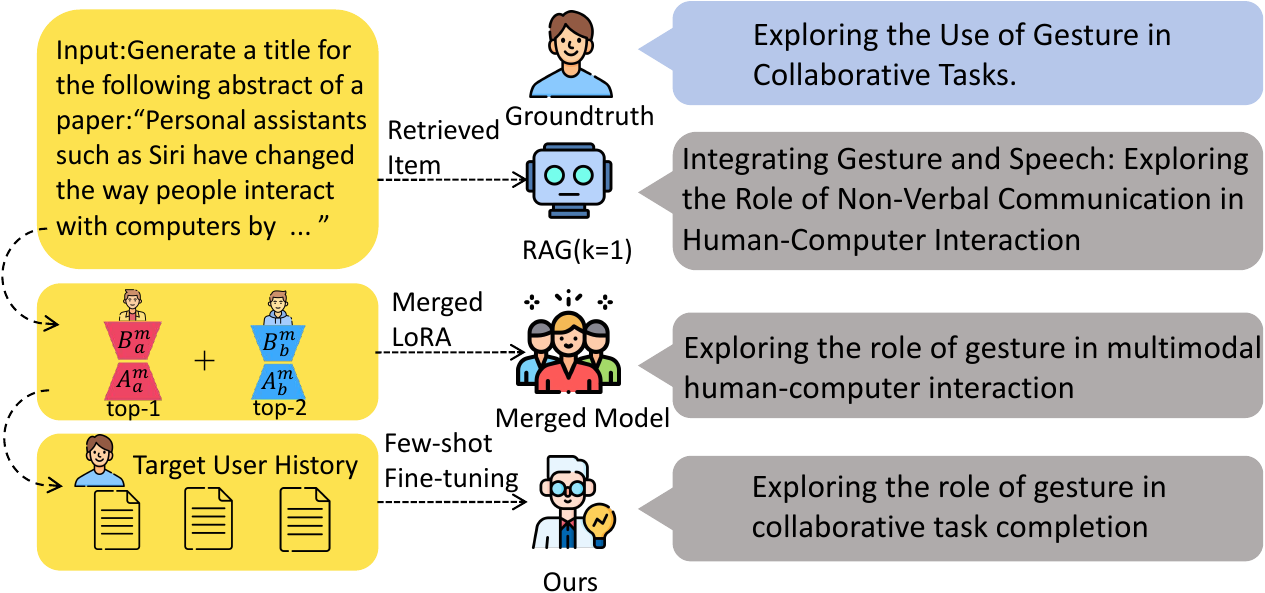} 
    \caption{Case study on personalized title generation (LaMP-5) for a user specializing in Computer Vision. Our full framework generates the most accurate title by capturing the user's task-oriented focus.} 
    \label{fig:case_study}
\end{figure}
\subsection{Case Study}
Figure~\ref{fig:case_study} presents a case study on the LaMP-5 task to demonstrate our framework's effectiveness. The target user (`user\_id: 11001393') is an expert in Computer Vision, with a publication history focused on geometric matching, image segmentation, and object recognition. The query asks for a title for an abstract about using gestures in collaborative tasks for human-computer interaction.
The RAG (k=1) baseline produces a verbose title focused on ``Speech-Human-Computer Interaction'', missing the key concept of ``Collaborative Tasks''. The Merged-Only Model, benefiting from the merged LoRAs of similar users, shows improvement by identifying the general ``multimodal human-computer interaction'' theme but still lacks specificity. 

In contrast, our complete framework (Ours) generates a highly precise title, ``Exploring the role of gesture in collaborative task completion'', which is closely match the ground truth, demonstrating the critical importance of our final adaptation stage. Our framework successfully distills the user's specific, task-oriented research focus (evident from their history on ``model matching'' and ``task completion'') and then effectively applies it to the new domain. This targeted approach enables the system to correctly identify and emphasize the ``collaborative task'' component that other models overlook, resulting in superior performance.
\section{Related Work}
LLMs excel at general knowledge tasks but struggle with user-specific personalization. Current personalized LLM research follows two main approaches: prompt-based and fine-tuning-based methods. Prompt-based methods~\cite{mysore2023pearl, richardson2023integrating, liu2024once, li2024matryoshka} directly incorporate personalized information into prompts, offering flexibility and cost-effectiveness but suffering from noise susceptibility and privacy risks. Fine-tuning-based methods typically employ Parameter-Efficient Fine-Tuning (PEFT) for fine-tuning, privacy-preserving personalization through internal parameters. These include shared LoRA~\cite{hu2022lora} modules for cross-user generalization, user-specific approaches like PLoRA~\cite{zhang2024personalized} and LM-P~\cite{wozniak2024personalized} that incorporate user embeddings and IDs respectively, and one-LoRA-per-user paradigms for enhanced personalization. Notable implementations include OPPU~\cite{tan2024democratizing}, which combines private PEFT with runtime profile augmentation, PerPCS~\cite{tan2024personalized}, employing dynamic LoRA routing via gating networks and PROPER~\cite{zhang2025proper}, which progressively aligns personalization across population, group, and user levels. Our MTA advances beyond static one-LoRA-per-user designs by constructing a meta-LoRA bank with dynamic merging, supplemented by LoRA stacking for finer-grained alignment, achieving accurate personalized adaptation even with limited samples.
\section{Conclusion}
In this paper, we introduced MTA, a novel three-stage framework for PLLMs centered on a ``Merge-then-Adapt'' strategy to address the critical challenges of scalability and limited adaptability inherent in existing one-LoRA-per-user models. MTA first constructs a Meta-LoRA Bank from diverse anchor users. It then employs an adaptive merging technique to create a customized foundation for target users, followed by a final, ultra-low-rank LoRA stacking adaptation step to capture fine-grained preferences. Our comprehensive experiments on the LaMP benchmark demonstrate that MTA significantly outperforms prompt-based and prominent fine-tuning baselines. Furthermore, MTA's training time and parameter storage costs are also significantly reduced. Through its dynamic, collaborative paradigm, MTA makes it possible to build scalable and efficient personalized LLMs.



\bibliography{aaai2026}
\appendix       
\section*{Appendix}

\section{Task Details} 

Here we present a detailed breakdown of the seven personalization tasks. The objective of these descriptions is to provide a clear understanding of the format and requirements for each task.

\begin{itemize}[leftmargin=*]
    \item \textbf{Personalized Citation Identification:} This is a binary text classification task. Given a paper with title $x$ authored by user $u$, and two candidate papers for citation, the model must predict which of the two candidates user $u$ is more likely to cite. The prediction is based on the user's history, which consists of the titles and abstracts of their previous publications.

    \item \textbf{Personalized Movie Tagging:} As a 15-way text classification task, the goal here is to predict a tag for a movie that aligns with a user's personal tagging preferences. The input to the model is a movie description $x$. The model must output a suitable tag, chosen from a set of fifteen, based on the user's historical movie-tag associations.

    \item \textbf{Personalized Product Rating:} This task can be framed as either a 5-way text classification or a regression problem. The model receives a product review $x$ written by user $u$. Its objective is to predict the integer rating (from 1 to 5) that user $u$ would assign to the product. This prediction is guided by the user's past review and rating pairs.

    \item \textbf{Personalized News Headline Generation:} This is a text generation task designed to assess a model's capacity to emulate an author's distinct style. The input query $x$ is the body of a news article. The model is required to generate a headline $y$ for this article that reflects the stylistic patterns observed in the user's historical article-headline pairs.

    \item \textbf{Personalized Scholarly Title Generation:} Similar to the news headline task, this is a text generation challenge set in an academic context. Given an abstract of a research paper $x$, the model's task is to generate a title $y$. The generated title should be stylistically consistent with the author's previous work, as evidenced by their historical abstract-title pairs.
\end{itemize}

We exclude the \textbf{LaMP-6: Email Subject Generation} task due to restricted access to its private dataset.

Furthermore, we choose to exclude \textbf{LaMP-7: Personalized Tweet Paraphrasing} because its data structure presents a fundamental conflict with our framework's design. In LaMP-7, the provided user history is a holistic corpus of a user's past tweets, rather than a collection of structured query-answer pairs. Specifically, the history contains an author's original tweets, while the task is to paraphrase a new, normalized sentence in that author's style. This format prevents us from constructing a training dataset directly from the user's history.

Benchmark methods such as \textbf{OPPU} and \textbf{PROPER} adopt a multi-stage process to handle such scenarios. Their process begins with a \textbf{population-level adaptation} stage. In this initial step, the model is fine-tuned on a general task dataset using query data from a pool of non-target users, with the goal of learning broad, task-related capabilities before any personalization occurs. To simplify the overall process and improve efficiency, our framework is designed to circumvent this large-scale, preparatory pre-training. Therefore, we omit the LaMP-7 task from our experiments.

\section{Datasets Details}

Table \ref{tab:dataset_stats} presents the statistics for the number of user history items across the different LaMP tasks~\cite{salemi2023lamp}. As stated in the main text, our work focuses on evaluating personalization in data-scarce scenarios. Consistent with this focus, we curate our test sets by selecting users who possess a limited number of historical interactions. Specifically, for each task, we randomly select 100 users for the test set based on the length of their history. For the LaMP-2 and LaMP-4 tasks, we choose users with 10 history items each. For LaMP-1 and LaMP-5, users with 50 history items are selected. Finally, for the LaMP-3 task, we select users with 100 history items.

\begin{table}[h!]
\centering
\caption{Statistics on the number of history items per user for each LaMP task.}
\label{tab:dataset_stats}
\begin{tabular}{lccc}
\toprule
\textbf{Task} & \textbf{Max Items} & \textbf{Min Items} & \textbf{Avg. Items} \\
\midrule
LaMP-1 & 607 & 40 & 88.4 \\
LaMP-2 & 1289 & 4 & 204.46 \\
LaMP-3 & 1021 & 97 & 202.52 \\
LaMP-4 & 945 & 7 & 31.2 \\
LaMP-5 & 911 & 47 & 94.34 \\
\bottomrule
\end{tabular}
\end{table}

\section{Baseline Details}

We compare our proposed framework against two main categories of personalization baselines: prompt-based and fine-tuning-based methods. Further details on each are provided below.

\begin{itemize}[leftmargin=*]
    \item \textbf{RAG}: This is a prompt-based personalization method introduced in the LaMP benchmark~\cite{salemi2023lamp}. For a given user query, this approach retrieves the top-$k$ most relevant items from that user's historical data corpus. These retrieved items are then concatenated with the original query to form an augmented prompt, which is subsequently fed to the LLM to generate a personalized response. In our experiments, we evaluate RAG with the number of retrieved items $k$ set to 1, 3, and 5, and compare our framework against its strongest performance across these settings.

    \item \textbf{OPPU}~\cite{tan2024democratizing}: This fine-tuning-based framework, proposed by~, pioneers the ``one-PEFT-per-user'' paradigm. It addresses challenges of model ownership and generalization by training a dedicated, lightweight PEFT module (specifically LoRA) for each individual user from scratch, using that user's entire behavior history. The resulting personal PEFT module encapsulates the user's specific preferences and can be "plugged into" a base LLM to create a personalized model. This approach requires separate training and storage for each user, leading to computation and storage costs that scale linearly with the user base.

    \item \textbf{PER-PCS}~\cite{tan2024personalized}: This framework introduces a more efficient, collaborative approach to PEFT-based personalization. Instead of training a new PEFT module for every target user, PER-PCS first trains PEFTs for a select group of representative "sharer" users. These PEFTs are then broken down into smaller modules, or "pieces". For a new target user, PER-PCS assembles a personalized PEFT in a training-free manner by selecting and combining the most relevant pieces from the shared pool, guided by the target user's own history data. This method achieves performance comparable to OPPU but with significantly lower computation and storage requirements.

    \item \textbf{PROPER}~\cite{zhang2025proper}: This framework addresses the data scarcity issue in personalization through a progressive, three-stage learning process. It introduces a meso-level, group-based adaptation between the population and user levels. The process is as follows: (1) \textbf{Population-Level Adaptation}, where a base LoRA is trained on a general user pool for task alignment; (2) \textbf{Group-Level Adaptation}, which uses a Mixture-of-Experts (MoE) structure to learn shared preferences for automatically clustered user groups; and (3) \textbf{User-Level Adaptation}, where a final, user-specific LoRA is trained to capture the remaining individual preferences. This hierarchical approach aims to provide a better starting point for personalizing data-sparse users.
\end{itemize}

\section{Hyperparameter Details}

The specific hyperparameter settings used for the final adaptation stage of our framework are detailed in Table \ref{tab:my_hyperparams_pivoted_compact}. We note that the LoRA rank, per device train batch size, and gradient accumulation steps are kept consistent across all tasks, while the number of training epochs and the learning rate are varied to optimize performance for each specific task.

\begin{table}[h!]
\centering
\caption{Hyperparameter settings for our framework across the evaluated LaMP tasks.}
\label{tab:my_hyperparams_pivoted_compact}
\begin{tabular}{lccccc}
\toprule
\textbf{Task} & \textbf{Rank} & \textbf{Epochs} & \textbf{LR} & \textbf{BS} & \textbf{Accum.} \\
\midrule
LaMP-1        & 4             & 2               & $5 \times 10^{-5}$   & 2           & 4                 \\
LaMP-2        & 4             & 3               & $1 \times 10^{-4}$   & 2           & 4                 \\
LaMP-3        & 4             & 1               & $5 \times 10^{-5}$   & 2           & 4                 \\
LaMP-4        & 4             & 3               & $1 \times 10^{-4}$   & 2           & 4                 \\
LaMP-5        & 4             & 2               & $5 \times 10^{-5}$   & 2           & 4                 \\
\bottomrule
\end{tabular}
\end{table}

\section{Prompt Details}

We present the prompt templates used for each task in our experiments. The text in curly braces, such as \texttt{\{USER HISTORY\}}, is replaced with content specific to different users and queries.

\begin{tcolorbox}[
  colback=gray!10,
  colframe=gray,
  width=\linewidth,
  arc=1mm,
  auto outer arc,
  title={LaMP-1: Personalized Citation Identification},
  breakable,
]
\noindent \#\#\# User History: \\
\texttt{\{USER HISTORY\}} \\[1ex]
\noindent \#\#\# User Instruction: \\
For an author who has written the paper with the title "\texttt{\{QUERY PAPER TITLE\}}", which reference is related? Just answer with [1] or [2] without explanation. \\
Reference: [1] - \texttt{\{OPTION 1\}} [2] - \texttt{\{OPTION 2\}} \\
Answer:
\end{tcolorbox}

\begin{tcolorbox}[
  colback=gray!10,
  colframe=gray,
  width=\linewidth,
  arc=1mm,
  auto outer arc,
  title={LaMP-2: Personalized Movie Tagging},
  breakable,
]
\noindent \#\#\# User History: \\
\texttt{\{USER HISTORY\}} \\[1ex]
\noindent \#\#\# User Instruction: \\
Which tag does this movie relate to among the following tags? Just answer with the tag name without further explanation. \\
tags: [sci-fi, based on a book, comedy, action, twist ending, dystopia, dark comedy, classic, psychology, fantasy, romance, thought-provoking, social commentary, violence, true story] \\
description: \texttt{\{QUERY MOVIE DESCRIPTION\}} \\
Tag:
\end{tcolorbox}

\begin{tcolorbox}[
  colback=gray!10,
  colframe=gray,
  width=\linewidth,
  arc=1mm,
  auto outer arc,
  title={LaMP-3: Personalized Product Rating},
  breakable,
]
\noindent \#\#\# User History: \\
\texttt{\{USER HISTORY\}} \\[1ex]
\noindent \#\#\# User Instruction: \\
What is the score of the following review on a scale of 1 to 5? just answer with 1, 2, 3, 4, or 5 without further explanation. \\
review: \texttt{\{QUERY REVIEW\}} \\
Score:
\end{tcolorbox}

\begin{tcolorbox}[
  colback=gray!10,
  colframe=gray,
  width=\linewidth,
  arc=1mm,
  auto outer arc,
  title={LaMP-4: Personalized News Headline Generation},
  breakable,
]
\noindent \#\#\# User History: \\
\texttt{\{USER HISTORY\}} \\[1ex]
\noindent \#\#\# User Instruction: \\
Generate a headline for the following article. \\
Article: \texttt{\{QUERY ARTICLE\}} \\
Headline:
\end{tcolorbox}

\begin{tcolorbox}[
  colback=gray!10,
  colframe=gray,
  width=\linewidth,
  arc=1mm,
  auto outer arc,
  title={LaMP-5: Personalized Scholarly Title Generation},
  breakable,
]
\noindent \#\#\# User History: \\
\texttt{\{USER HISTORY\}} \\[1ex]
\noindent \#\#\# User Instruction: \\
Generate a title for the following abstract of a paper. \\
Abstract: \texttt{\{QUERY ABSTRACT\}} \\
Title:
\end{tcolorbox}

\section{Supplementary Experiment}
To analyze the impact of the rank ($r$) of the final adaptation LoRA on performance, we conduct a supplementary experiment. We evaluate the performance of our full framework on two representative tasks (LAMP-3 Product Rating and LAMP-5 Scholarly Title Generation) across a range of different ranks ($r \in \{2, 4, 6, 8, 16\}$).

\begin{figure}[h!]
    \centering
    \begin{subfigure}[b]{0.48\columnwidth}
        \centering
        \includegraphics[width=\textwidth]{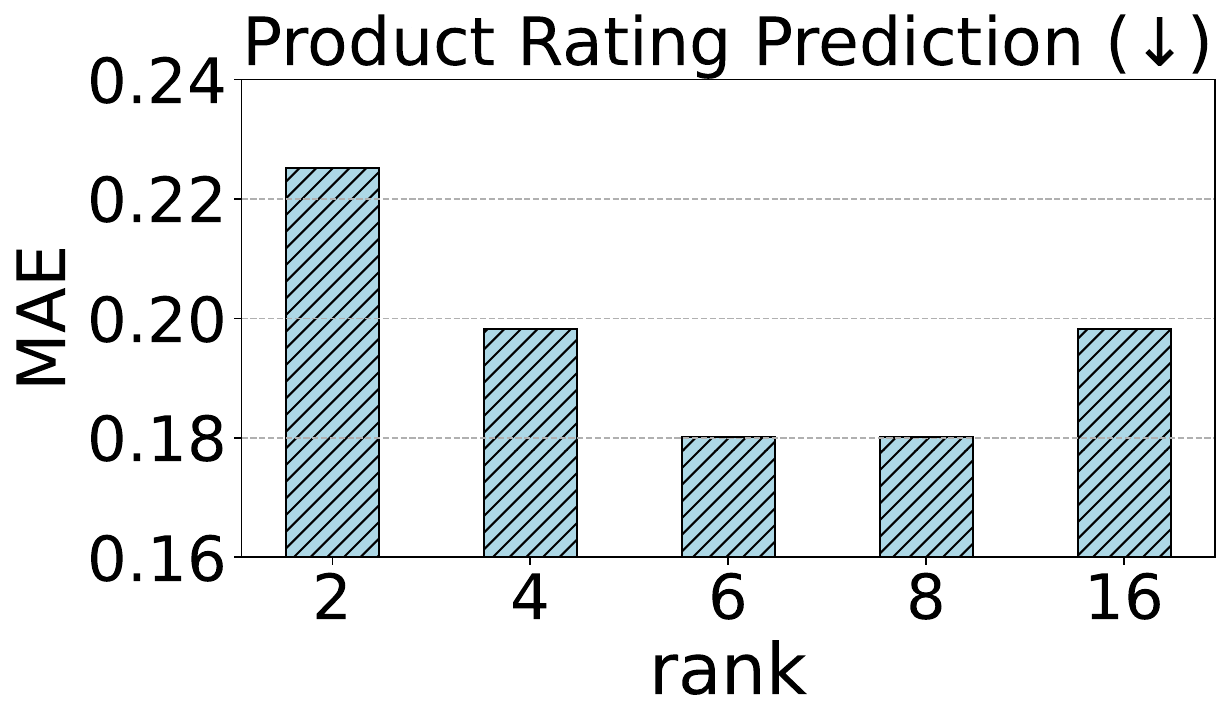}
    \end{subfigure}
    \hfill 
    \begin{subfigure}[b]{0.48\columnwidth}
        \centering
        \includegraphics[width=\textwidth]{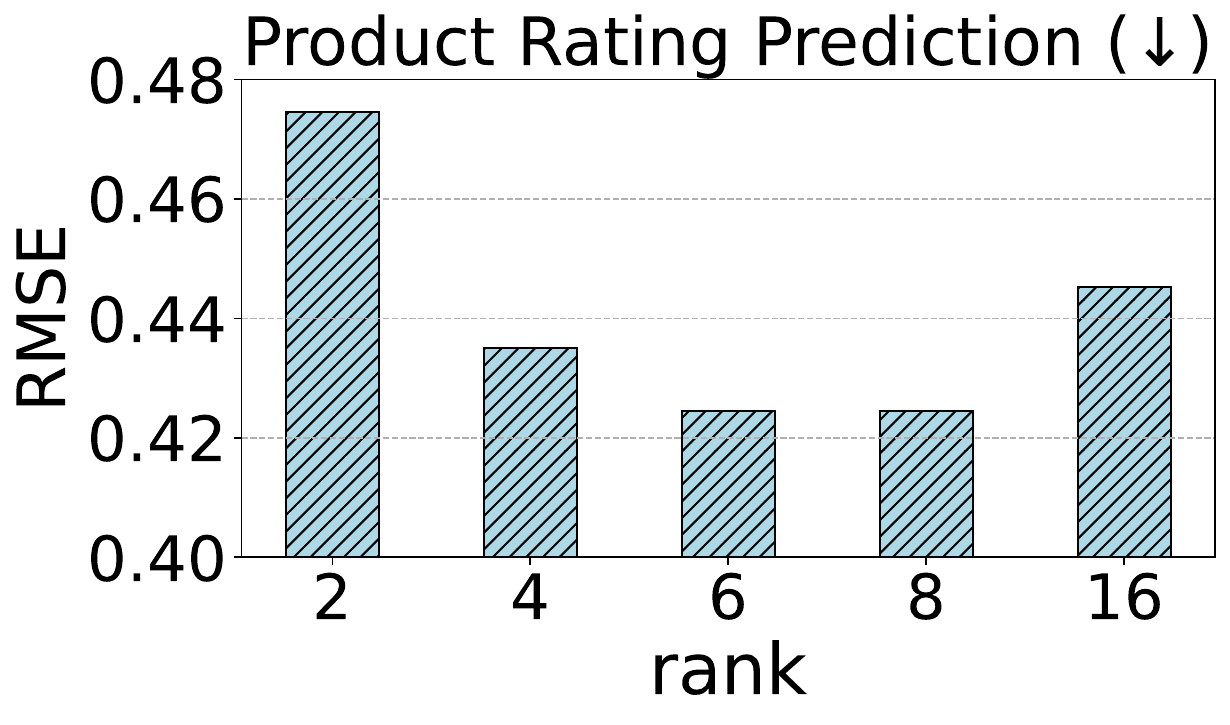}
    \end{subfigure}

    \vspace{0.3cm} 

    \begin{subfigure}[b]{0.48\columnwidth}
        \centering
        \includegraphics[width=\textwidth]{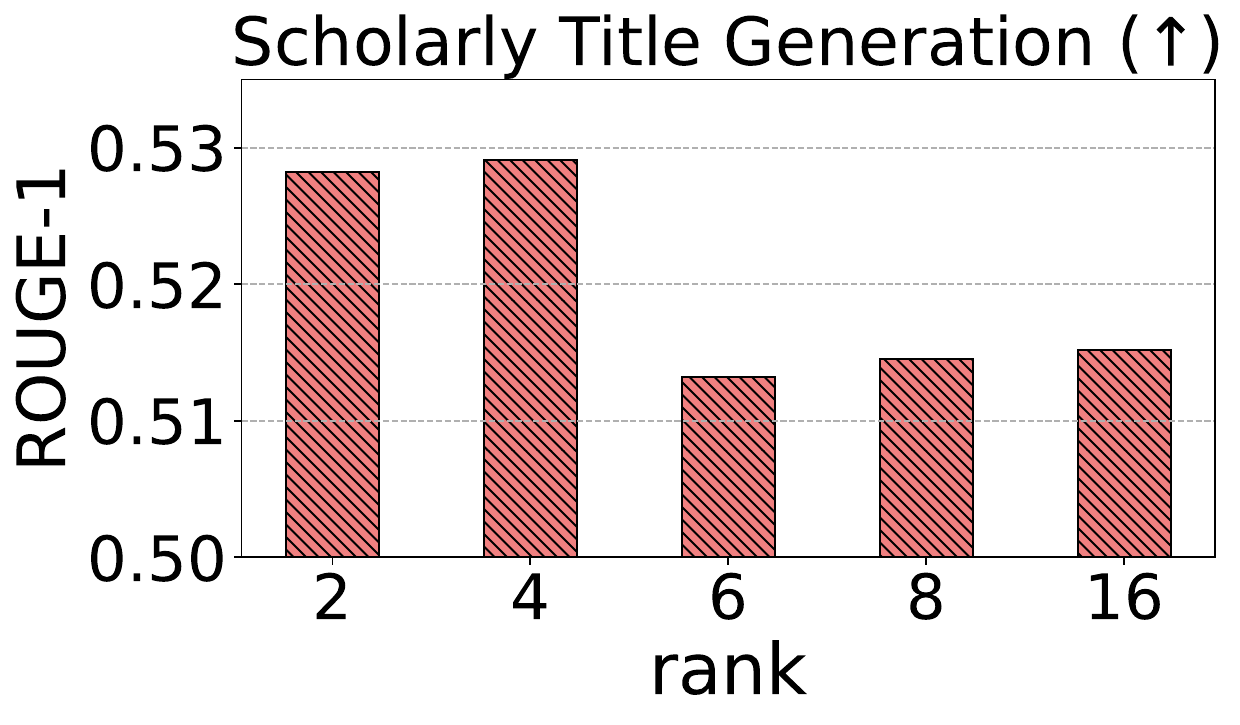}
    \end{subfigure}
    \hfill 
    \begin{subfigure}[b]{0.48\columnwidth}
        \centering
        \includegraphics[width=\textwidth]{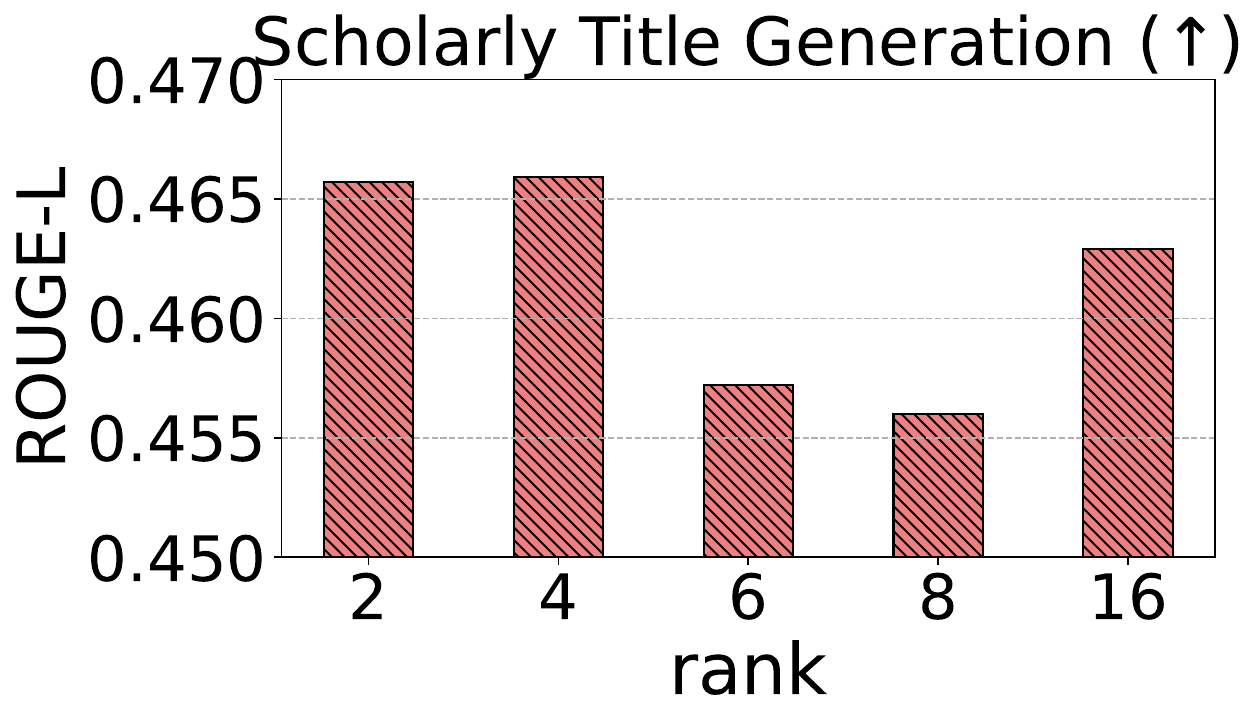}
    \end{subfigure}

    \caption{Performance analysis of the final adaptation LoRA with varying ranks ($r$). The top row shows Product Rating (LAMP-3) results; the bottom row shows Scholarly Title Generation (LAMP-5) results.}
    \label{fig:rank_ablation}
\end{figure}

The results are shown in Figure~\ref{fig:rank_ablation}. We observe that the optimal rank is task-dependent. For the Product Rating task, the model achieves the best performance on both MAE and RMSE metrics at a rank of 6 or 8. In contrast, the optimal rank for the Scholarly Title Generation task is 4. Notably, the relationship between performance and rank is non-monotonic; for both tasks, performance degrades when the rank is increased to 16.

This finding confirms that using an extremely low rank is an efficient and effective strategy. Our analysis shows that selecting a small rank (such as $r=4$) achieves optimal or near-optimal performance while minimizing the parameter count. This validates our choice of $r=4$ as the default setting in our main experiments, striking an excellent balance between performance and efficiency.

\end{document}